\documentclass{article}
\usepackage{spconf,amsmath,graphicx,hyperref}
\usepackage{amsmath}
\usepackage{multirow}
\usepackage{booktabs}
\usepackage{amssymb}

\title{TEXTS-Diff: TEXTS-Aware Diffusion Model for Real-World Text Image Super-Resolution}
\name{Haodong He$^*$, Xin Zhan$^*$, Yancheng Bai$^\ddagger$, Rui Lan, Lei Sun, Xiangxiang Chu
\thanks{\raggedright $^*$ Equal contribution \\ \hspace{13pt}  
$^\ddagger$ Project leader: Yancheng Bai \\ \hspace{13pt}
Email: haodonghe@whu.edu.cn, xinzhan@hust.edu.cn, \{yancheng.byc, lr264907, ally.sl, chuxiangxiang.cxx\}@alibaba-inc.com
}
\address{Amap, Alibaba Group}
}

\begin{document}
\ninept
\maketitle
\begin{abstract}
Real-world text image super-resolution aims to restore overall visual quality and text legibility in images suffering from diverse degradations and text distortions.
However, the scarcity of text image data in existing datasets results in poor performance on text regions.
In addition, datasets consisting of isolated text samples limit the quality of background reconstruction.
To address these limitations, we construct Real-Texts, a large-scale, high-quality dataset collected from real-world images, which covers diverse scenarios and contains natural text instances in both Chinese and English.
Additionally, we propose the TEXTS-Aware Diffusion Model (TEXTS-Diff) to achieve high-quality generation in both background and textual regions.
This approach leverages abstract concepts to improve the understanding of textual elements within visual scenes and concrete text regions to enhance textual details.
It mitigates distortions and hallucination artifacts commonly observed in text regions, while preserving high-quality visual scene fidelity.
Extensive experiments demonstrate that our method achieves state-of-the-art performance across multiple evaluation metrics, exhibiting superior generalization ability and text restoration accuracy in complex scenarios. All the code, model, and dataset will be released.
\end{abstract}
\section{Introduction}
\label{sec:intro}

Real-world image super-resolution (Real-ISR) \cite{RealESRGAN2021} aims to reconstruct high-resolution (HR) contents from low-resolution (LR) images degraded by complex real-world factors. The derived task of real-world text image super-resolution (Real-TISR) encounters an even tougher scenario: it requires not only the reconstruction of global scene details to ensure visual coherence, but also the precise restoration of character structures in text regions.
This necessitates that the model needs to simultaneously preserve the super-resolution (SR) quality of both background and text regions.

However, existing SR benchmark datasets \cite{Lim_2017_CVPR_Workshops,li2023lsdir} contain a relatively small number of images with textual content, and no corresponding text annotations are provided.
Although some datasets \cite{wang2020scene,ma2023benchmark,singh2021textocr} provide annotated image-text pairs, limitations such as limited scale, isolated text, or monolingual composition constrain their practical applicability.
The absence of high-quality text image datasets for this task has limited the generalization of existing models, even though they exhibit remarkable performance in general image \cite{wang2024exploiting,yang2024pixel,yu2024scaling}.
They yield distorted glyphs, missing strokes, or hallucinated characters in the text super-resolution results, primarily due to the minimal attention paid to textual regions.
In addition, many dedicated text SR methods \cite{chen2021scene,zhang2024diffusion,zhu2023improving} enhance only the legibility of cropped text patches while ignoring the surrounding background, leading to significant performance degradation when applied to LR images from complex natural scenes.
Consequently, current models struggle to effectively balance textual and background regions, which poses a significant challenge for real-world applications.

\begin{figure}[!t]
\centering
\includegraphics[width=0.48\textwidth]{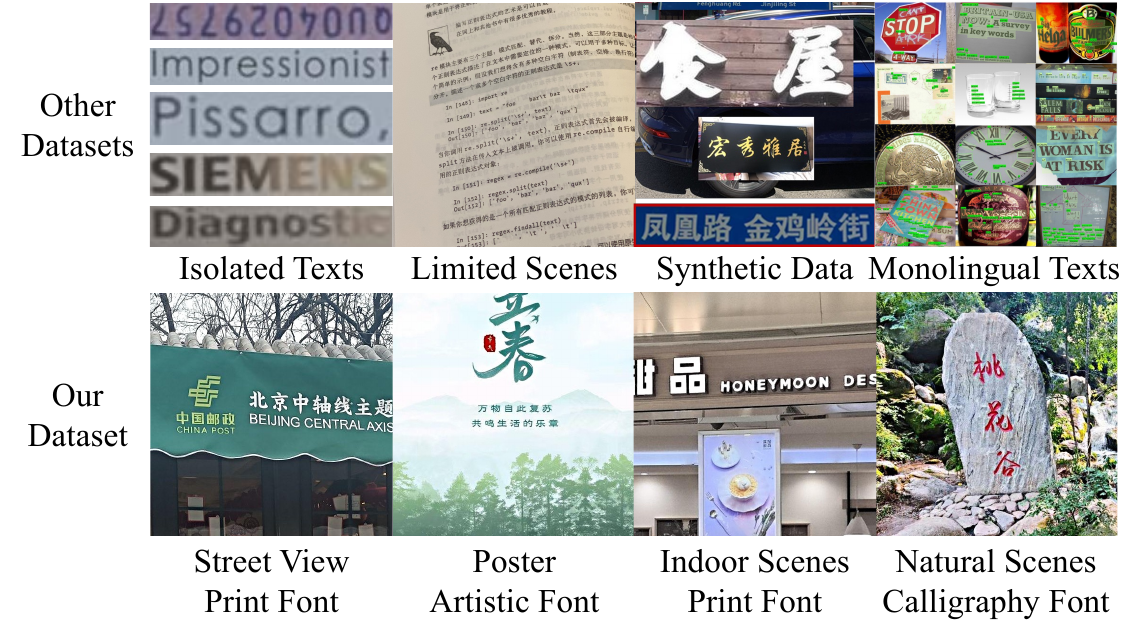}
\vspace{-0.8cm}
\caption{
Example images from other datasets and our Real-Texts dataset. Our dataset includes Chinese and English text across a wide range of scene types and font styles.
}
\vspace{-0.5cm}
\label{fig:dataset}
\end{figure}

To remedy these issues, we construct Real-Texts, a large-scale text image SR dataset. It contains more than 30,000 real-world images with bilingual text (Chinese and English), covering a variety of font styles and diverse scene types.
Meanwhile, to achieve high-quality generation in both background and textual regions, we draw inspiration from the global-to-local cognitive pattern in human reading, and propose TEXTS-Diff, a text-aware one-step diffusion model for the Real-TISR task. Specifically, we use the abstract concept of “TEXTS” as a semantic anchor within the primary visual features of LR images, guiding these visual cues to establish text-aware conceptual representations. We then fuse these visual cues with high-level features from the text detection model to further enhance their perception of concrete text regions. Finally, conditioned on the guidance of these visual cues, the diffusion model performs one-step image SR, significantly improving text legibility, while maintaining the ability to super-resolve the entire image and achieve fast inference.

Our contributions are: 1)  We construct Real-Texts, a text SR image dataset. It contains more than 30,000 Chinese-English images captured under diverse scenes and font styles. 2) We introduce TEXTS-Diff, which incorporates both abstract text concepts and concrete text regions into visual cues to guide the diffusion process, thereby enhancing the restoration of textual structures in images. 3)  Extensive experiments demonstrate that our approach surpasses current methods on dedicated text image SR benchmarks and also delivers competitive performance on general SR datasets.

\section{Real-Texts Dataset}
Real-TISR task requires a dataset that includes images containing textual content together with their corresponding bounding boxes and text labels. As shown in Table \ref{tab:dataset}, although general SR datasets such as RealSR \cite{realsr}, Flickr2K \cite{Lim_2017_CVPR_Workshops}, and LSDIR \cite{li2023lsdir} provide high-quality images, they contain a limited number of images with text and provide no associated annotations. 
TextZoom \cite{wang2020scene} annotates existing SR datasets \cite{realsr,zhang2019zoom}, but its scale and diversity are somewhat limited.
TextOCR \cite{singh2021textocr} is a large-scale SR dataset (containing 903,069 words).
Although it has diverse font styles and scene types, it contains only English text, which limits its linguistic richness.
In contrast, Real-CE \cite{ma2023benchmark} contains images with both Chinese and English text, but its scale is limited.
Recent work \cite{hu2025tadisr} synthesizes data by overlaying text onto images.
It leads to a noticeable gap from natural images, because the pasted text is unrelated to the background.
To address these limitations, we construct the Real-Texts dataset.

\subsection{Dataset Creation Process}
We collect a set of real-world images with varying resolutions.
Then, we apply PPOCRv5 \cite{cui2025paddleocr30technicalreport} for text detection and recognition, removing any images without valid OCR results.
Each retained image (a total of 72,476) is subsequently cropped into regions of 512 × 512 pixels according to the detected text bounding boxes.
In this way, a total of 74,035 cropped images are obtained. These cropped images are then assessed with the VisualQuality-R1 model \cite{wu2025visualquality} to obtain quality scores. 
After automatically filtering the images by a preset threshold of 4.25, we further manually verified the results to ensure high quality.
Finally, following the degradation pipeline of Real-ESRGAN \cite{RealESRGAN2021}, we generate the LR counterparts.
Through these processes, 34,875 pairs of images are eventually obtained.

\begin{table}[!t]
\centering
\setlength{\tabcolsep}{1.5pt}
\renewcommand{\arraystretch}{0.9}
\begin{tabular}{|l|c|c|c|c|c|c|c|}
\hline
\textbf{Dataset} & \textbf{Text} & \textbf{Bilingual} & \textbf{Font} & \textbf{Scene} & \textbf{\#Img} & \textbf{Text line} \\
\hline
RealSR \cite{realsr}    & $\times$ & $\times$ & $\times$ & $\times$ & 559     & - \\
TextZoom  \cite{wang2020scene} & $\checkmark$ & $\times$ & $\times$ & $\times$ & 1,059   & 21,740 \\
TextOCR \cite{singh2021textocr}   & $\checkmark$ & $\times$ & $\checkmark$ & $\checkmark$ & 28,134  & - \\
Real-CE \cite{ma2023benchmark}  & $\checkmark$ & $\checkmark$ & $\times$ & $\checkmark$ & 2,718   & 33,789 \\
\textbf{Real-Texts} & $\checkmark$ & $\checkmark$ & $\checkmark$ & $\checkmark$ & \textbf{34,875}  & \textbf{136,136} \\
\hline
\end{tabular}
\vspace{-0.2cm}
\caption{Comparison between Real-Texts and other datasets.}
\vspace{-0.4cm}
\label{tab:dataset}
\end{table}

\subsection{Dataset Analysis}
Real-Texts contains 34,875 LR–HR image pairs with both Chinese and English text, covering various scenarios and font styles. As shown in Figure \ref{fig:dataset}, the fonts encompass various styles, including printed, artistic, and calligraphic fonts, while the image scenes feature diverse types such as street views (6,124), posters (5,910), indoor scenes (19,223), and natural scenes (3,618). The dataset contains a total of 136,136 textual instances, including 75,962 Chinese, 37,971 English, and 3,476 mixed Chinese and English instances, while the remaining 18,727 correspond to texts or symbols from other languages. We assign 33,875 image pairs to the training dataset and the remaining 1,000 pairs to the test dataset.

\section{Methodology}
\label{sec:Methodology}
\subsection{Overview Architecture}
\begin{figure}[!t]
\centering
\includegraphics[width=0.5\textwidth]{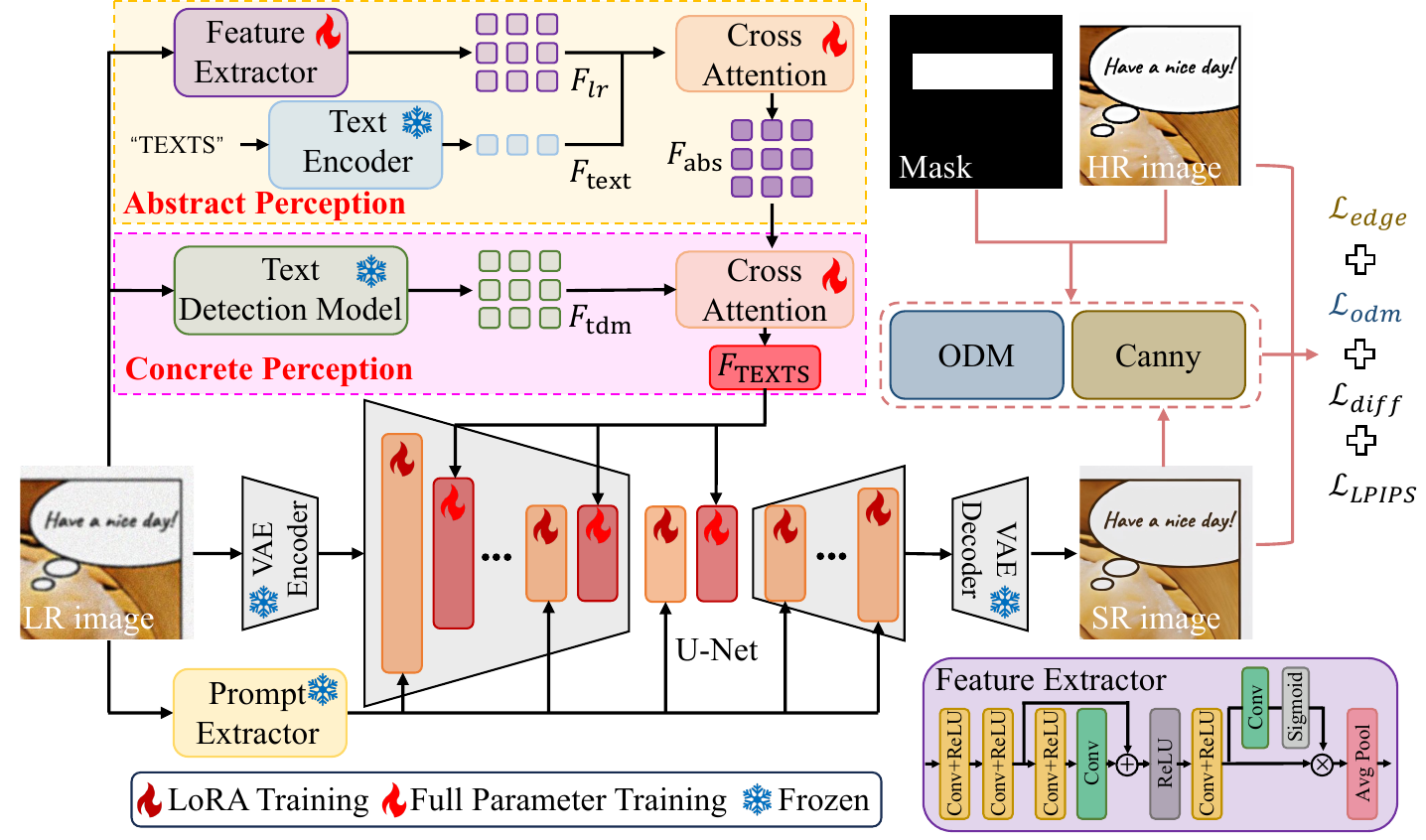} 
\vspace{-0.8cm}
\caption{
Overview of TEXTS-Diff. The framework captures text-aware visual cues to guide the diffusion process through abstract and concrete perception processes.
}
\label{fig:framework}
\vspace{-0.4cm}
\end{figure}

Current general diffusion-based SR methods \cite{seesr2024,yu2024scaling,diffbir2024,chen2025faithdiff} exhibit insufficient text awareness, resulting in poor text legibility. Additionally, most of these methods rely on multi-step generation of HR images, which leads to slow inference. To address these limitations, we propose TEXTS-Diff, a one-step diffusion model for Real-TISR. Its overall architecture is illustrated in Figure \ref{fig:framework}. Inspired by the global-to-local cognitive pattern in human reading, TEXTS-Diff incorporates both abstract text concepts and concrete text regions into visual cues, which are then used as conditions to guide the model's diffusion process. 

\subsection{TEXTS-Aware Guidance}
Advanced diffusion-based SR models \cite{StableSR2024,seesr2024} enhance computational efficiency by performing the diffusion process in the latent space of noisy LR images. Nevertheless, the VAE used for latent encoding attenuates high-frequency visual details, which are indispensable for accurate reconstruction of HR images. This loss of information often yields over-smoothed SR outputs, thus degrading the sharpness and continuity of text structures.
To address this issue, we propose Feature Extractor ($\mathcal{FE}$) module (see the bottom-right of Figure \ref{fig:framework}) to obtain primary visual cues and compensate for information loss. Specifically, $\mathcal{FE}$ progressively extracts features through stacked convolutional layers and residual blocks. Given an LR image $ I_{lr} \in \mathbb{R}^{H \times W \times 3}$, the module produces visual features with high-frequency components better preserved, denoted as $F_{lr}$:
{
\setlength{\abovedisplayskip}{5pt}
\setlength{\belowdisplayskip}{5pt}
\begin{equation}
    F_{lr} = \mathcal{FE}(I_{lr}), \quad F_{lr} \in \mathbb{R}^{h \times w \times c},
\end{equation}
}
where $H \times W$ denotes the spatial resolution of the LR image, $h \times w$ refers to the spatial resolution of the LR latent features after $8 \times$ downsampling, and $c$ is the channel dimension.  

To enable the visual features to form an abstract conceptual understanding of text, we establish a process of textual abstraction perception by applying cross-attention between the visual features $F_{lr}$ and the ``TEXTS" textual features $F_{\text{text}}$. 
The abstract perception process can be expressed as:
{
\setlength{\abovedisplayskip}{5pt}
\setlength{\belowdisplayskip}{5pt}
\begin{equation}
    \begin{aligned}
        & F_{\text{text}} = \mathcal{CLIP}(''\text{TEXTS}''), \\
        & Q_{lr} = W^{lr}_q \cdot F_{lr},\quad K_a=W^{a}_k \cdot F_{\text{text}}, \quad V_a = W^{a}_v \cdot F_{\text{text}},\\
        & F_{\text{abs}} = \text{softmax}\left(\frac{Q_{lr} \cdot K_a^T}{\sqrt{d}}\right) \cdot V_a,
    \end{aligned}
    \label{eq:abs}
\end{equation}
} where $\mathcal{CLIP}$ refers to the text encoder \cite{LDM2022}, and $W^{lr}_q$, $W^{a}_k$, and $W^{a}_v$ are learnable linear projection matrices. $Q_{lr}$, $K_a$, and $V_a$ denote the query, key, and value matrices, respectively. $d$ indicates the feature dimension, whereas $F_{\text{abs}}$ represents the visual feature enhanced by abstract text concepts. 
In order to enhance the perception of $F_{\text{abs}}$ with respect to concrete text regions, we construct a concrete perception process by applying cross-attention between $F_{\text{abs}}$ and the high-level visual features $F_{\text{tdm}}$ extracted from the original LR image by the text detection model ($\mathcal{TDM}$). The  process can be defined as:
{
\setlength{\abovedisplayskip}{5pt}
\setlength{\belowdisplayskip}{5pt}
\begin{equation}
    \begin{aligned}
        & F_{\text{tdm}} = \mathcal{TDM}(I_{lr}), \\
        & Q_{\text{abs}} = W^{\text{abs}}_q \cdot F_{\text{abs}},\quad K_c = W^{c}_k \cdot F_{\text{tdm}}, \quad V_c = W^{c}_v \cdot F_{\text{tdm}},\\
        & F_{\text{TEXTS}} = \text{softmax}\left(\frac{Q_{\text{abs}} \cdot K_c^T}{\sqrt{d}}\right) \cdot V_c,
    \end{aligned}
\end{equation}
}
where $\mathcal{TDM}$ denotes DBNet++ \cite{liao2022dbnetpp}, and $W^{\text{abs}}_q$, $W^{c}_k$, and $W^{c}_v$ are learnable linear matrices. $Q_{\text{abs}}$, $K_c$, and $V_c$ denote the query, key, and value matrices, respectively. Through these interaction processes from abstract to concrete, we obtain text-aware enhanced visual features $F_{\text{TEXTS}}$ to guide SR.

\begin{table*}[!t]
\centering
\small
\setlength{\tabcolsep}{1.5pt}
\renewcommand{\arraystretch}{0.9}
\begin{tabular}{c|l|c|ccccccccc}
\toprule
Datasets & Methods & OCR-A $\uparrow$ & PSNR $\uparrow$ & SSIM $\uparrow$ & LPIPS $\downarrow$ & DISTS $\downarrow$ & FID $\downarrow$ & NIQE $\downarrow$ & MANIQA $\uparrow$ & MUSIQ $\uparrow$ & CLIPIQA $\uparrow$ \\
\midrule

\multirow{8}{*}{\begin{tabular}[c]{@{}c@{}}\textit{Real-CE}\end{tabular}}
& Real-ESRGAN \cite{RealESRGAN2021}& 0.3829 & 19.10 & 0.6529 & 0.3962 & 0.2460 & 54.66 & 5.3908 & 0.4250 & 56.56 & 0.6421 \\
& StableSR \cite{StableSR2024} (4 steps) & \underline{0.3938} & 19.28 & 0.6876 & 0.3472 & 0.1866 & 50.04 & 4.9530 & 0.4748 & 49.74 & 0.5247 \\
& DiffBIR \cite{diffbir2024} (50 steps) & 0.3697 & 19.96 & 0.6711 & 0.3585 & 0.2251 & 49.25 & 5.1390 & 0.4990 & 56.29 & 0.5935\\
& FaithDiff \cite{chen2025faithdiff} (20 steps) & 0.3691 & 19.54 & 0.6569 & 0.3337 & 0.1726 & \textbf{42.95} & \textbf{4.1747} & \textbf{0.6189} & \underline{64.28} & \underline{0.6619} \\
& SeeSR \cite{seesr2024} (50 steps) & 0.3846 & 19.86 & 0.7078 & 0.3254 & 0.2163 & 48.34 & \underline{4.6469} & 0.5770 & \textbf{64.98} & \textbf{0.6884} \\
& SUPIR \cite{supir} (50 steps) & 0.3590 & \underline{20.00} & 0.6584 & 0.3537 & 0.1852 & 44.52 & 4.6995 & 0.5344 & 51.19 & 0.5525 \\
& OSEDiff \cite{wu2024osediff} (1 step) & 0.3758 & 19.47 & \underline{0.7097} & \underline{0.3123} & \underline{0.1645} & 49.71 & 5.2332 & \underline{0.5933}  & 63.38 & 0.6439 \\
& \textbf{TEXTS-Diff} (1 step) & \textbf{0.4631} & \textbf{20.31} & \textbf{0.7235} & \textbf{0.2791} & \textbf{0.1414} & \underline{44.36} & 5.4176 & 0.5590 & 56.10 & 0.5439 \\
\midrule

\multirow{8}{*}{\begin{tabular}[c]{@{}c@{}}\textit{Real-Texts}\end{tabular}}
& Real-ESRGAN \cite{RealESRGAN2021} & 0.3764 & 22.91 & 0.7265 & 0.2794 & 0.1878 & 50.12 & 6.5263 & 0.5645 & 67.99 & 0.6009 \\
& StableSR \cite{StableSR2024} (4 steps) & 0.3511 & \underline{23.69} & \underline{0.7481} & 0.2396 & 0.1502 & 36.23 & 4.8865 & 0.5867 & 57.08 & 0.4707 \\
& DiffBIR \cite{diffbir2024} (50 steps) & 0.3538 & 22.63 & 0.6712 & 0.2709 & 0.1632 & 41.76 & 4.9049 & \underline{0.6744} & \underline{72.33} & \underline{0.6501} \\
& FaithDiff \cite{chen2025faithdiff} (20 steps) & 0.3412 & 22.82 & 0.6979 &\underline{ 0.2188} & \underline{0.1448} & 34.21 & \textbf{4.3931} & \textbf{0.6942} & 71.43 & 0.6313 \\
& SeeSR \cite{seesr2024} (50 steps) & 0.3229 & 23.09 & 0.7151 & 0.2347 & 0.1750 & 41.88 & 5.0474 & 0.6742 & \textbf{72.69} & \textbf{0.6882} \\
& SUPIR \cite{supir} (50 steps) & \underline{0.3811} & 23.23 & 0.6960 & 0.2453 & 0.1454 & \underline{31.73} & 4.9894 & 0.6579 & 68.94 & 0.6409 \\
& OSEDiff \cite{wu2024osediff} (1 step) & 0.3049 & 22.91 & 0.7160 & 0.2291 & 0.1476 & 38.21 & 4.6753 & 0.6600  & 70.12 & 0.6171 \\
& \textbf{TEXTS-Diff} (1 step) & \textbf{0.3951} & \textbf{24.49} & \textbf{0.7499} & \textbf{0.1801} & \textbf{0.1174} & \textbf{27.70} & \underline{4.5296} & 0.6482 & 66.18 & 0.5258 \\
\midrule
\multirow{8}{*}{\begin{tabular}[c]{@{}c@{}}\textit{RealSR}\end{tabular}}
& Real-ESRGAN \cite{RealESRGAN2021} & - & \underline{25.69} & \textbf{0.7616} & \underline{0.2727} & 0.2063 & 135.18 & 5.8295 & 0.5487 & 60.18 & 0.4449 \\
& StableSR \cite{StableSR2024} (4 steps)& - & 24.70 & 0.7085 & 0.3018 & 0.2135 & 128.51 & 5.9122 & 0.6221 & 65.78 & 0.6178 \\
& DiffBIR \cite{diffbir2024} (50 steps) & - & 24.77 & 0.6572 & 0.3658 & 0.2310 & 128.99 & 5.5696 & 0.6253 & 64.85 & 0.6386 \\
& FaithDiff \cite{chen2025faithdiff} (20 steps) & - & 25.32 & 0.7091 & 0.2807 & \underline{0.1922} & \textbf{108.68} & \textbf{4.9688} & \textbf{0.6635} & 67.37 & 0.5942 \\
& SeeSR \cite{seesr2024} (50 steps) & - & 25.18 & 0.7216 & 0.3009 & 0.2223 & 125.55 & 5.4081 & \underline{0.6442} & \textbf{69.77} & \underline{0.6612} \\
& SUPIR \cite{supir} (50 steps) & - & 25.16 & 0.6945 & 0.3336 & 0.2102 & 124.00 & \underline{5.2011} & 0.5714 & 56.94 & 0.5206 \\
& OSEDiff \cite{wu2024osediff} (1 step) & - & 25.15 & 0.7341 & 0.2921 & 0.2128 & 123.49 & 5.6476 & 0.6326 & \underline{69.09} & \textbf{0.6693} \\
& \textbf{TEXTS-Diff} (1 step) & - & \textbf{26.51} & \underline{0.7602} & \textbf{0.2485} & \textbf{0.1842} & \underline{109.11} & 5.5610 & 0.6154 & 64.82 & 0.6042 \\
\bottomrule
\end{tabular}
\vspace{-0.2cm}
\caption{
Quantitative comparison with state-of-the-art methods on the benchmarks. The best and second-best results of each metric are highlighted in bold and underline, respectively.
}
\label{tab:results}
\vspace{-0.4cm}
\end{table*}

\subsection{Diffusion Processing}
We employ a pretrained Stable Diffusion (SD) model as the prior. For practical SR applications, we follow the one-step diffusion technique of OSEDiff \cite{mmsr2025} by directly transition from the LR image distribution to the HR image distribution in one step. To accommodate this modification, the SD U-Net is fine-tuned using LoRA \cite{LoRA}. To guide the diffusion process, two conditional features are injected into the U-Net: the text-aware feature $F_{\text{TEXTS}}$, and the semantic visual feature $F_p$ extracted from the LR image by a pretrained Prompt Extractor ($\mathcal{PE}$) \cite{seesr2024}. Specifically, $F_p$ is fed into the original Transformer blocks, whereas $F_{\text{TEXTS}}$ is supplied to newly added trainable Transformer blocks inserted after each down block and the middle block of the U-Net. 
Under this guidance, the U-Net refines the latent feature $F_v$ produced by the VAE Encoder and reconstructs the SR image $I_{sr}$ through the VAE Decoder. 
It can be described as:
{\setlength{\abovedisplayskip}{5pt}
\setlength{\belowdisplayskip}{5pt}
\begin{equation}
    \begin{aligned}
         &F_v = Encoder(I_{lr}), \quad F_p = \mathcal{PE}(I_{lr}), \\
         &I_{sr} = Decoder(\text{U-Net}(F_{\text{TEXTS}}, F_p, F_v)). \\ 
    \end{aligned}
\end{equation}
}
\subsection{Loss Function}
To train our SR framework, we impose an L2-norm loss between the ground truth HR images $I_{hr}$ and the SR predictions $I_{sr}$ to steer the diffusion process. In addition, we incorporate an LPIPS \cite{lpips} perceptual loss on HR-SR image pairs to encourage the reconstructed images to be perceptually closer to the ground truth. These losses are formulated as follows:
{\setlength{\abovedisplayskip}{5pt}
\setlength{\belowdisplayskip}{5pt}
\begin{equation}
    \mathcal{L}_{\text{diff}} = \|I_{sr} - I_{hr}\|_2 , \quad
    \mathcal{L}_{\text{LPIPS}} = \text{LPIPS}(I_{sr}, I_{hr}).
\end{equation}
}
To enhance text reconstruction in SR images, we use the Canny edge detector to extract edges from SR and HR images, and apply an L2-based edge loss $\mathcal{L}_{\text{edge}}$ for text regions. The OCR-Text Destylization Modeling method ($\mathcal{ODM}$) \cite{duan2024odm} is also used to define $\mathcal{L}_{\text{odm}}$, improving the model's sensitivity to textual differences. The losses are formulated as follows:
{
\setlength{\abovedisplayskip}{5pt}
\setlength{\belowdisplayskip}{5pt}
\begin{equation}
    \begin{aligned}
        &\mathcal{L}_{\text{edge}} = \|(\text{Canny}(I_{sr}) - \text{Canny}(I_{hr})) \circ I_{mask}\|_2 , \\
        &\mathcal{L}_{\text{odm}} = \|(\mathcal{ODM}(I_{sr})-\mathcal{ODM}(I_{hr})) \circ I_{mask})\|_2,
    \end{aligned}
\end{equation} 
}
where $\circ$ denotes the pixel-wise multiplication, and $I_{mask}$ is the binarized ground truth mask based on text line annotations of the HR image. The total loss is formulated as a weighted combination of the above components using $\lambda_1$-$\lambda_4$ to balance different aspects of the reconstruction:
{
\setlength{\abovedisplayskip}{5pt}
\setlength{\belowdisplayskip}{5pt}
\begin{equation}
    \mathcal{L}_{\text{total}} = \lambda_1 \mathcal{L}_{\text{diff}} + \lambda_2 \mathcal{L}_{\text{LPIPS}} + \lambda_3 \mathcal{L}_{\text{edge}} + \lambda_4 \mathcal{L}_{\text{odm}},   
\end{equation}
}
where $\lambda_1$-$\lambda_4$ are set to 1, 2, 1, and 10, respectively.

\section{Experiments}
\subsection{Experimental Settings}
\subsubsection{Dataset and Details}

The training dataset composition is as follows: 33,875 images with text annotations from the Real-Texts, 14,312 cropped images with text annotations from the Real-CE training set \cite{ma2023benchmark}, and 20,000 images without text annotations from the LSDIR dataset \cite{li2023lsdir}.
We apply the widely used Real-ESRGAN \cite{RealESRGAN2021} for degradations. We evaluate model performance on the test sets of Real-CE and Real-Texts.
For implementation details, we utilize the pretrained SD2.1 model \cite{LDM2022} as the backbone. During training, the VAE encoder and decoder are frozen, and the U-Net is fine-tuned via LoRA \cite{LoRA}. 

\subsubsection{Evaluation Metrics}
We use OCR accuracy (OCR-A) to evaluate the effectiveness of text SR. Specifically, we evaluate the SR images with text annotations by cropping them based on the ground-truth (GT) text bounding boxes. 
Subsequently, we compare the text recognition results obtained from the PPOCRv5 \cite{cui2025paddleocr30technicalreport} with the GT text. To ensure reliability, we manually verify 100 random samples and confirm a text content accuracy of 94.2$\%$.
Completely identical text lines are counted as correct, whereas mismatched lines are considered errors. In addition, we use reference-based metrics such as PSNR, SSIM \cite{ssim}, LPIPS \cite{lpips}, DISTS \cite{dists}, and FID \cite{fid}, as well as no-reference metrics such as NIQE \cite{niqe}, MANIQA \cite{maniqa}, MUSIQ \cite{musiq}, and CLIPIQA \cite{clipiqa} to evaluate the model's SR performance.

\begin{figure}[!t]
\centering
\includegraphics[width=0.5\textwidth]{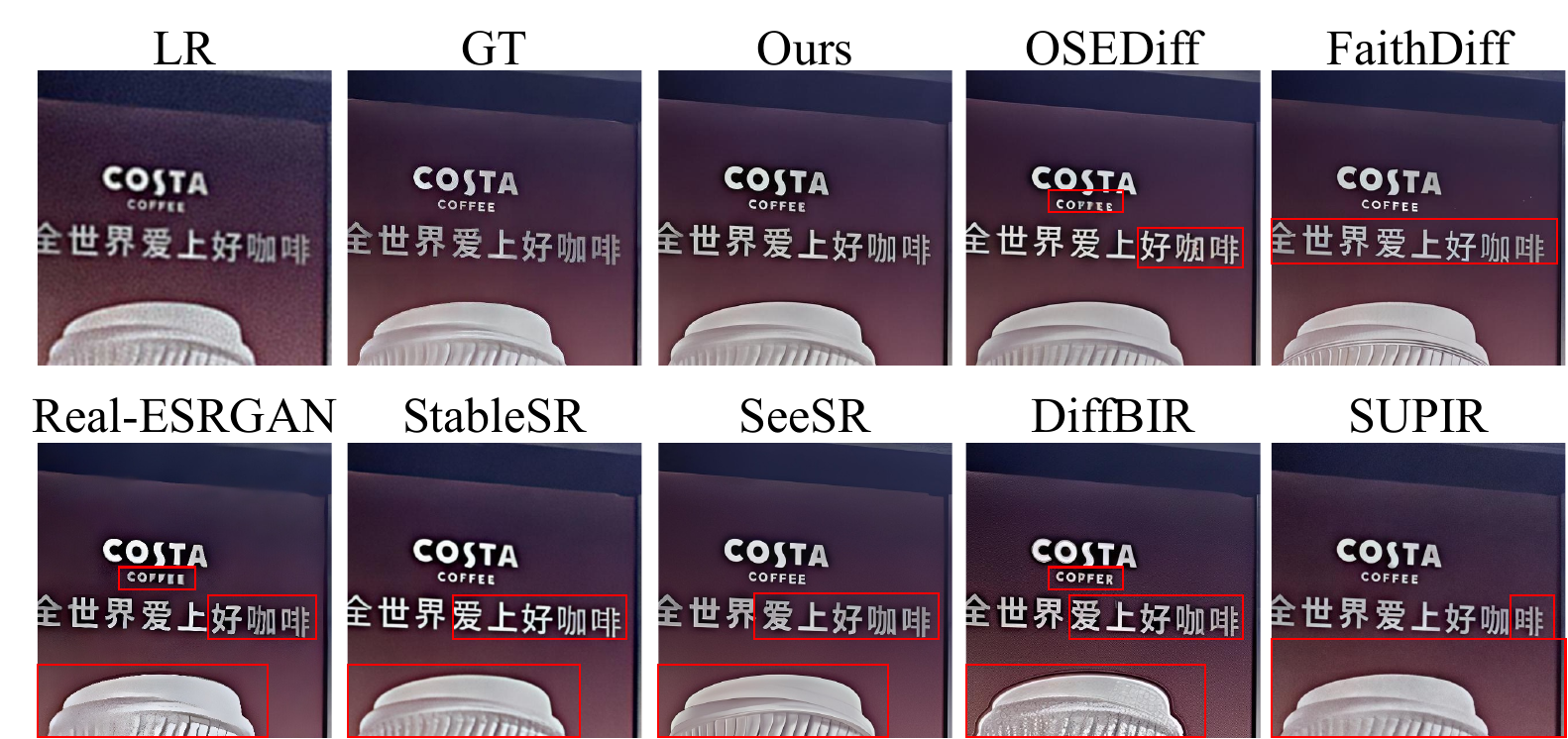}
\vspace{-0.6cm}
\caption{
Visual comparison on the Real-Texts test dataset. 
The red box means the poor result areas.
}
\label{fig:comparison our dataset}
\vspace{-0.4cm}
\end{figure}

\begin{figure}[!t]
\centering
\includegraphics[width=0.5\textwidth]{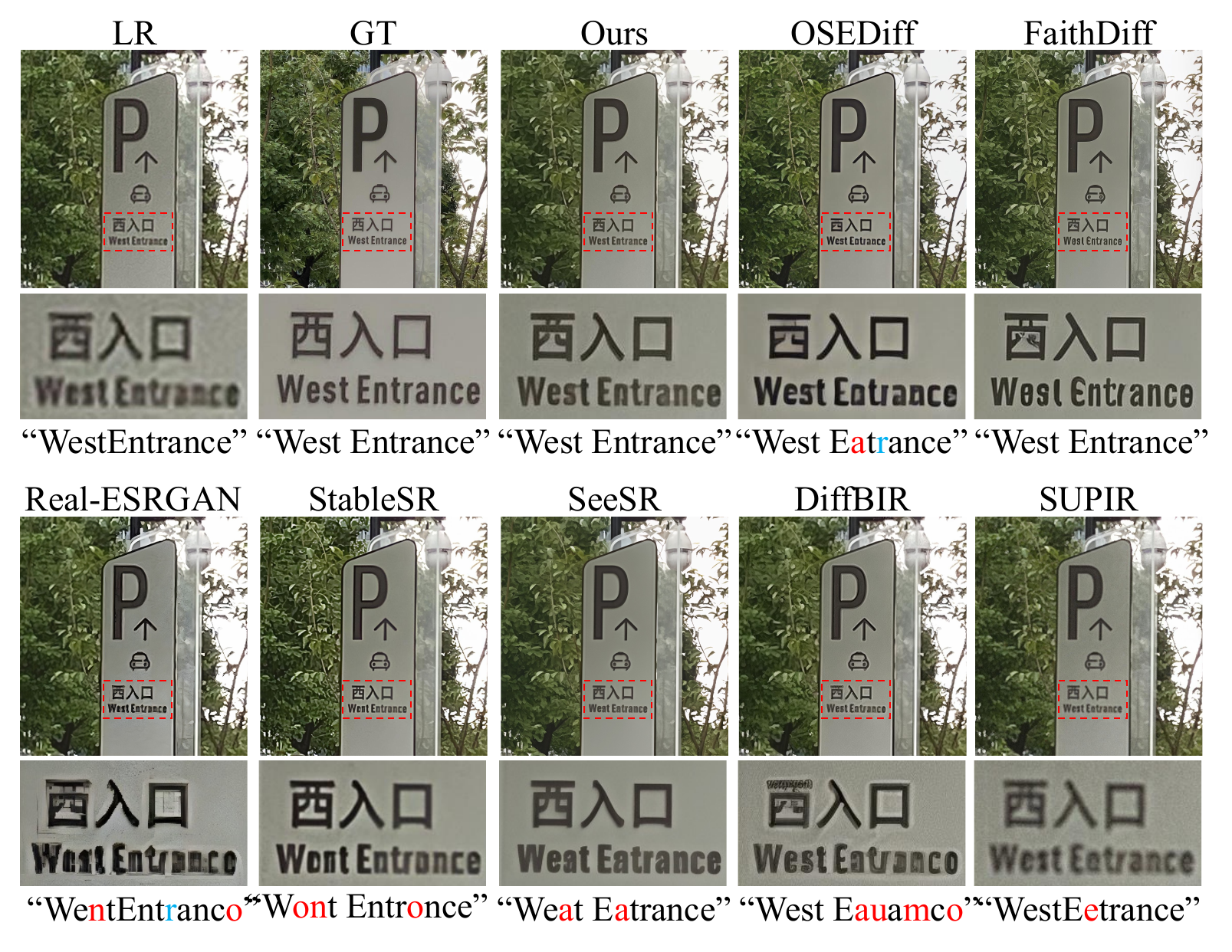} 
\vspace{-0.8cm}
\caption{
Visual comparison on the Real-CE validation dataset.
The enlarged part of the dashed red box is shown under the original results.
Their corresponding OCR recognition results in the aforementioned figure, where erroneous text is highlighted in red and omitted text is marked in blue.
}
\label{fig:comparison Real-CE}
\vspace{-0.4cm}
\end{figure}
\subsection{Experimental Results}

\subsubsection{Quantitative Comparison}
As shown in Table \ref{tab:results}, our model achieves state-of-the-art (SOTA) performance on the two annotated datasets (Real-Texts and Real-CE) for text region SR, as measured by OCR-A.
Our model's OCR-A surpasses that of the second-best model by 3.67\% and 17.60\% on Real-Texts and Real-CE, respectively.
By comprehensively analyzing both reference-based and non-reference metrics, we achieve the highest total number of best and second-best results on Real-Texts and Real-CE.
In addition, the result of RealSR dataset demonstrates that our approach restores text regions without compromising overall image SR quality.

\subsubsection{Qualitative Comparison}
As shown in the comparison results in Figures \ref{fig:comparison our dataset} and \ref{fig:comparison Real-CE}, our model consistently outperforms other baseline methods in restoring text regions. Specifically, our SR results significantly improve text legibility for both Chinese and English texts, while avoiding deformation and distortion. Moreover, the model maintains strong performance on non-text regions, effectively preserving overall image quality. These results demonstrate the effectiveness of our model, indicating its potential for practical applications in text image restoration.

\begin{table}[!h]
\centering
\small
\setlength{\tabcolsep}{1.5pt}
\renewcommand{\arraystretch}{0.9}
\begin{tabular}{c|cccc}
\toprule
Metrics & Ours & w/o TDM & w/o TEXTS & w/o Both\\
\midrule
OCR-A $\uparrow$ & \textbf{0.3951} & 0.3764 & \underline{0.3829} & 0.3715 \\
PSNR $\uparrow$ & \underline{24.49} & 24.46 & 24.44 & \textbf{24.52} \\
SSIM $\uparrow$ & \underline{0.7499} & 0.7487 & 0.7480 & \textbf{0.7503} \\
LPIPS $\downarrow$ & \textbf{0.1801} & \underline{0.1807} & 0.1827 & 0.1826 \\
DISTS $\downarrow$ & \textbf{0.1174} & \underline{0.1179} & 0.1192 & 0.1180 \\
FID $\downarrow$ & \textbf{27.70} & \underline{27.77} & 28.08 & 27.88 \\
NIQE $\downarrow$ & \underline{4.5296} & \textbf{4.5159} & 4.6032 & 4.6619 \\
MANIQA $\uparrow$ & 0.6482 & \underline{0.6514} & \textbf{0.6521} & 0.6443 \\
MUSIQ $\uparrow$ & 66.18 & \underline{66.71} & \textbf{67.27} & 66.44 \\
CLIPIQA $\uparrow$ & 0.5258 & \underline{0.5358} & \textbf{0.5421} & 0.5282 \\
\bottomrule
\end{tabular}
\vspace{-0.2cm}
\caption{
Ablation study on Real-Texts.
}
\label{table:ablation}
\vspace{-0.4cm}
\end{table}
\subsubsection{Ablation Study}
We conduct experiments on the Real-Texts dataset to evaluate the effectiveness of the ``TEXTS'' concept and TDM features.
The results in Table \ref{table:ablation} show that incorporating both abstract and concrete perceptions of text in visual cues can significantly enhance text perception and restoration, while minimally affecting fidelity (PSNR -0.03 dB) and perceptual quality (NIQE +0.0137).

\section{Conclusion}
To address the challenges faced in the real-world text image super-resolution task, we propose Real-Texts, a large-scale, high-quality super-resolution dataset collected from real-world images. In addition, we present TEXTS-Diff, which introduces abstract text concepts and concrete text regions into visual guidance cues and enables one-step generation. This design significantly enhances text super-resolution performance while maintaining its capability to super-resolve image backgrounds.

\bibliographystyle{IEEEbib}
\bibliography{strings,refs}

\end{document}